\documentclass[10pt,twocolumn,letterpaper]{article}

\usepackage{cvpr}
\usepackage{tabu}
\usepackage{times}
\usepackage{epsfig}
\usepackage{graphicx}
\usepackage{amsmath}
\usepackage{amssymb}
\usepackage{adjustbox}
\usepackage{array}
\usepackage{subcaption}
\usepackage{verbatim}
\usepackage{balance}
\usepackage{booktabs}
\usepackage{multirow}

\newcommand{\ra}[1]{\renewcommand{\arraystretch}{#1}}

\newcommand\ver[1]{\rotatebox{90}{#1}}
\newcommand\minisection[1]{\vspace{1mm}\noindent \textbf{#1}}

\def\aa{\mathbf a}
\def\bb{\mathbf b}
\def\ff{\mathbf f}
\def\gg{\mathbf g}
\def\pp{\mathbf p}

\def\CC{\mathrm C}
\def\HH{\mathrm H}
\def\WW{\mathrm W}

\def\B{\mathcal B}

\def\G{\mathcal G}

\DeclareMathOperator*{\argmin}{arg\,min}

\newcommand{\timess}{\mathbin{\!\times\!}}

\usepackage[pagebackref=true,breaklinks=true,colorlinks,bookmarks=false]{hyperref}

\cvprfinalcopy 

\def\cvprPaperID{153} 

\ifcvprfinal\fi
\begin{document}

\title{Dilated Residual Networks}

\author{Fisher Yu \\
Princeton University\\
\and 
Vladlen Koltun \\
Intel Labs
\and 
Thomas Funkhouser \\
Princeton University
\\
}

\maketitle

\begin{abstract}
Convolutional networks for image classification progressively reduce resolution until the image is represented by tiny feature maps in which the spatial structure of the scene is no longer discernible. Such loss of spatial acuity can limit image classification accuracy and complicate the transfer of the model to downstream applications that require detailed scene understanding. These problems can be alleviated by dilation, which increases the resolution of output feature maps without reducing the receptive field of individual neurons. We show that dilated residual networks (DRNs) outperform their non-dilated counterparts in image classification without increasing the model's depth or complexity. We then study gridding artifacts introduced by dilation, develop an approach to removing these artifacts (`degridding'), and show that this further increases the performance of DRNs. In addition, we show that the accuracy advantage of DRNs is further magnified in downstream applications such as object localization and semantic segmentation.
\end{abstract}

\section{Introduction}

Convolutional networks were originally developed for classifying hand-written digits~\cite{LeCun1989}. More recently, convolutional network architectures have evolved to classify much more complex images~\cite{alexnet,vggnet,googlenet,resnet}.
Yet a central aspect of network architecture has remained largely in place. Convolutional networks for image classification progressively reduce resolution until the image is represented by tiny feature maps that retain little spatial information ($7\timess 7$ is typical).

While convolutional networks have done well, the almost complete elimination of spatial acuity may be preventing these models from achieving even higher accuracy, for example by preserving the contribution of small and thin objects that may be important for correctly understanding the image. Such preservation may not have been important in the context of hand-written digit classification, in which a single object dominated the image, but may help in the analysis of complex natural scenes where multiple objects and their relative configurations must be taken into account.

Furthermore, image classification is rarely a convolutional network's raison d'{\^e}tre. Image classification is most often a proxy task that is used to pretrain a model before it is transferred to other applications that involve more detailed scene understanding~\cite{rcnn,fcn}. In such tasks, severe loss of spatial acuity is a significant handicap. Existing techniques compensate for the lost resolution by introducing up-convolutions~\cite{fcn,Noh2015}, skip connections~\cite{Hariharan2015}, and other post-hoc measures.

Must convolutional networks crush the image in order to classify it? In this paper, we show that this is not necessary, or even desirable.  Starting with the residual network architecture, the current state of the art for image classification~\cite{resnet}, we increase the resolution of the network's output by replacing a subset of interior subsampling layers by dilation~\cite{dilation}.
We show that dilated residual networks (DRNs) yield improved image classification performance. Specifically, DRNs yield higher accuracy in ImageNet classification than their non-dilated counterparts, with no increase in depth or model complexity.

The output resolution of a DRN on typical ImageNet input is $28\timess 28$, comparable to small thumbnails that convey the structure of the image when examined by a human~\cite{Torralba2008}. While it may not be clear a priori that average pooling can properly handle such high-resolution output, we show that it can, yielding a notable accuracy gain. We then study gridding artifacts introduced by dilation, propose a scheme for removing these artifacts, and show that such `degridding' further improves the accuracy of DRNs.


We also show that DRNs yield improved accuracy on downstream applications such as weakly-supervised object localization and semantic segmentation. With a remarkably simple approach, involving no fine-tuning at all, we obtain state-of-the-art top-1 accuracy in weakly-supervised localization on ImageNet. We also study the performance of DRNs on semantic segmentation and show, for example, that a 42-layer DRN outperforms a ResNet-101 baseline on the Cityscapes dataset by more than 4 percentage points, despite lower depth by a factor of 2.4.

\section{Dilated Residual Networks}
\label{sec:drn}




Our key idea is to preserve spatial resolution in convolutional networks for image classification. Although progressive downsampling has been very successful in classifying digits or iconic views of objects, the loss of spatial information may be harmful for classifying natural images and can significantly hamper transfer to other tasks that involve spatially detailed image understanding. Natural images often feature many objects whose identities and relative configurations are important for understanding the scene. The classification task becomes difficult when a key object is not spatially dominant -- for example, when the labeled object is thin (e.g., a tripod) or when there is a big background object such as a mountain. In these cases, the background response may suppress the signal from the object of interest. What's worse, if the object's signal is lost due to downsampling, there is little hope to recover it during training. However, if we retain high spatial resolution throughout the model and provide output signals that densely cover the input field, backpropagation can learn to preserve important information about smaller and less salient objects.

The starting point of our construction is the set of network architectures presented by He et al.~\cite{resnet}. Each of these architectures consists of five groups of convolutional layers. The first layer in each group performs downsampling by striding: that is, the convolutional filter is only evaluated at even rows and columns. Let each group of layers be denoted by $\G^\ell$, for $\ell = 1,\ldots,5$. Denote the $i^{\text{th}}$ layer in group $\ell$ by $\G^\ell_i$. For simplicity of exposition, consider an idealized model in which each layer consists of a single feature map: the extension to multiple feature maps is straightforward. Let $f^\ell_i$ be the filter associated with layer $\G^\ell_i$. In the original model, the output of $\G^\ell_i$ is
\begin{equation}
(\G^\ell_i \ast f^\ell_i)(\pp) = \sum_{\aa + \bb = \pp} \G^\ell_i(\aa) \, f^\ell_i(\bb),
\end{equation}
where the domain of $\pp$ is the feature map in $\G^\ell_i$. This is followed by a nonlinearity, which does not affect the presented construction.

A naive approach to increasing resolution in higher layers of the network would be to simply remove subsampling (striding) from some of the interior layers. This does increase downstream resolution, but has a detrimental side effect that negates the benefits: removing subsampling correspondingly reduces the receptive field in subsequent layers. Thus removing striding such that the resolution of the output layer is increased by a factor of 4 also reduces the receptive field of each output unit by a factor of 4. This severely reduces the amount of context that can inform the prediction produced by each unit. Since contextual information is important in disambiguating local cues~\cite{GalleguillosBelongie2010}, such reduction in receptive field is an unacceptable price to pay for higher resolution. For this reason, we use dilated convolutions~\cite{dilation} to increase the receptive field of the higher layers, compensating for the reduction in receptive field induced by removing subsampling. The effect is that units in the dilated layers have the same receptive field as corresponding units in the original model.

\begin{figure}[t]
  \centering
    \begin{subfigure}[b]{0.9\linewidth}
        \includegraphics[width=\textwidth]{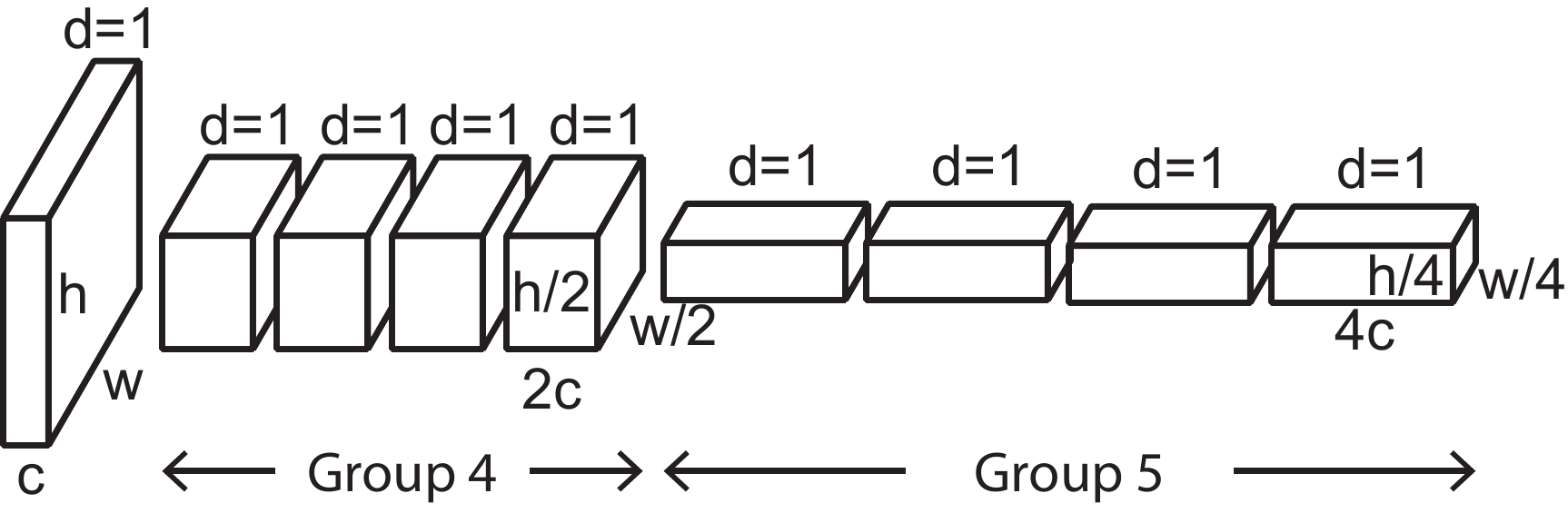}
        \caption{ResNet}
        \label{fig:convert_before}
    \end{subfigure}
    \vspace{3mm}\\
    \begin{subfigure}[b]{0.9\linewidth}
        \includegraphics[width=\textwidth]{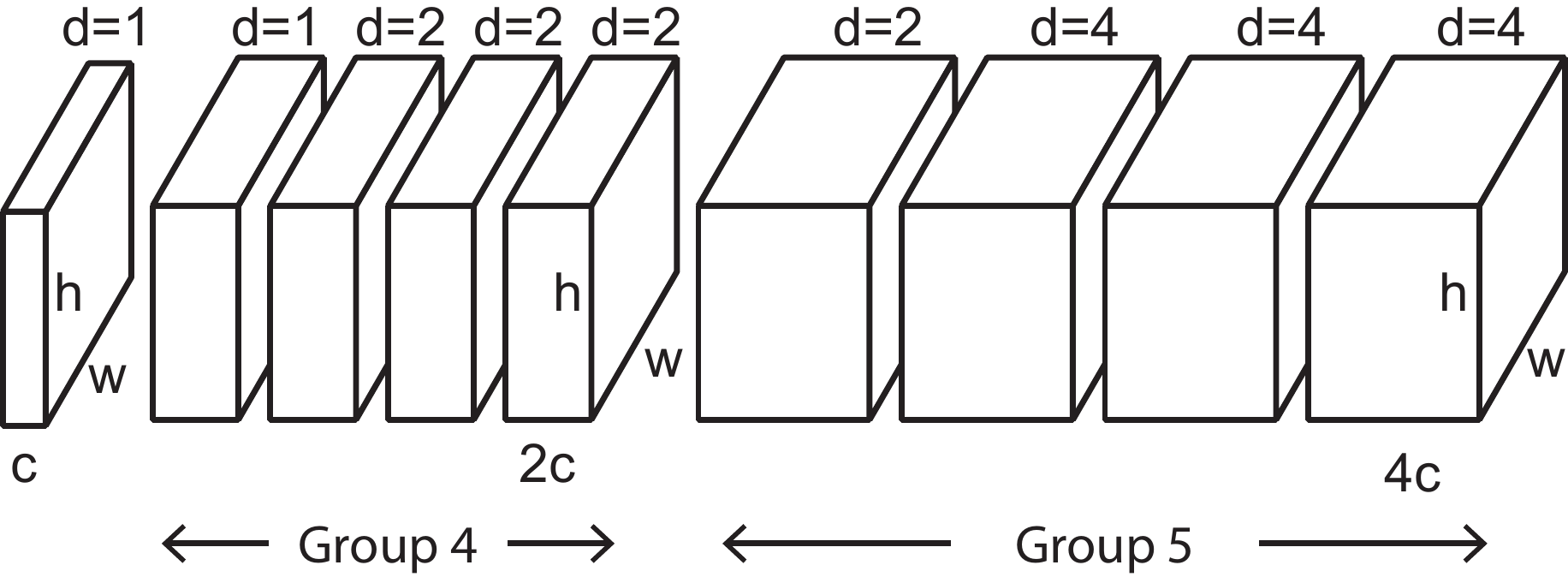}
        \caption{DRN}
        \label{fig:convert_after}
    \end{subfigure}
  \caption{Converting a ResNet into a DRN. The original ResNet is shown in (a), the resulting DRN is shown in (b). Striding in $\G^4_1$ and $\G^5_1$ is removed, bringing the resolution of all layers in $\G^4$ and $\G^5$ to the resolution of $\G^3$. To compensate for the consequent shrinkage of the receptive field, $\G^4_i$ and $\G^5_1$ are dilated by a factor of 2 and $\G^5_i$ are dilated by a factor of 4, for all $i\ge 2$. $c$, $2c$, and $4c$ denote the number of feature maps in a layer, $w$ and $h$ denote feature map resolution, and $d$ is the dilation factor.}
  \label{fig:convert}
\end{figure}

We focus on the two final groups of convolutional layers: $\G^4$ and $\G^5$. In the original ResNet, the first layer in each group ($\G^4_1$ and $\G^5_1$) is strided: the convolution is evaluated at even rows and columns, which reduces the output resolution of these layers by a factor of 2 in each dimension. The first step in the conversion to DRN is to remove the striding in both $\G^4_1$ and $\G^5_1$. Note that the receptive field of each unit in $\G^4_1$ remains unaffected: we just doubled the output resolution of $\G^4_1$ without affecting the receptive field of its units. However, subsequent layers are all affected: their receptive fields have been reduced by a factor of 2 in each dimension. We therefore replace the convolution operators in those layers by 2-dilated convolutions~\cite{dilation}:
\begin{equation}
(\G^4_i \ast_2 f^4_i)(\pp) = \sum_{\aa + 2\bb = \pp} \G^4_i(\aa) \, f^4_i(\bb)
\end{equation}
for all $i \ge 2$. The same transformation is applied to $\G^5_1$:
\begin{equation}
(\G^5_1 \ast_2 f^5_1)(\pp) = \sum_{\aa + 2\bb = \pp} \G^5_1(\aa) \, f^5_1(\bb).
\end{equation}
Subsequent layers in $\G^5$ follow two striding layers that have been eliminated. The elimination of striding has reduced their receptive fields by a factor of 4 in each dimension. Their convolutions need to be dilated by a factor of 4 to compensate for the loss:
\begin{equation}
(\G^5_i \ast_4 f^5_i)(\pp) = \sum_{\aa + 4\bb = \pp} \G^5_i(\aa) \, f^5_i(\bb)
\end{equation}
for all $i \ge 2$.
Finally, as in the original architecture, $\G^5$ is followed by global average pooling, which reduces the output feature maps to a vector, and a $1\timess 1$ convolution that maps this vector to a vector that comprises the prediction scores for all classes. The transformation of a ResNet into a DRN is illustrated in Figure~\ref{fig:convert}.


The converted DRN has the same number of layers and parameters as the original ResNet. The key difference is that the original ResNet downsamples the input image by a factor of 32 in each dimension (a thousand-fold reduction in area), while the DRN downsamples the input by a factor of 8. For example, when the input resolution is $224 \timess 224$, the output resolution of $\G^5$ in the original ResNet is $7 \timess 7$, which is not sufficient for the spatial structure of the input to be discernable. The output of $\G^5$ in a DRN is $28 \timess 28$. Global average pooling therefore takes in $2^4$ times more values, which can help the classifier recognize objects that cover a smaller number of pixels in the input image and take such objects into account in its prediction.


The presented construction could also be applied to earlier groups of layers ($\G^1$, $\G^2$, or $\G^3$), in the limit retaining the full resolution of the input. We chose not to do this because a downsampling factor of 8 is known to preserve most of the information necessary to correctly parse the original image at pixel level~\cite{fcn}. Furthermore, a $28\timess 28$ thumbnail, while small, is sufficiently resolved for humans to discern the structure of the scene~\cite{Torralba2008}. Additional increase in resolution has costs and should not be pursued without commensurate gains: when feature map resolution is increased by a factor of 2 in each dimension, the memory consumption of that feature map increases by a factor of 4. Operating at full resolution throughout, with no downsampling at all, is beyond the capabilities of current hardware.

\section{Localization}
\label{sec:vis}

Given a DRN trained for image classification, we can directly produce dense pixel-level class activation maps without any additional training or parameter tuning. This allows a DRN trained for image classification to be immediately used for object localization and segmentation.

\begin{figure}[t]
\centering
    \begin{subfigure}[b]{0.7\linewidth}
        \includegraphics[width=\textwidth]{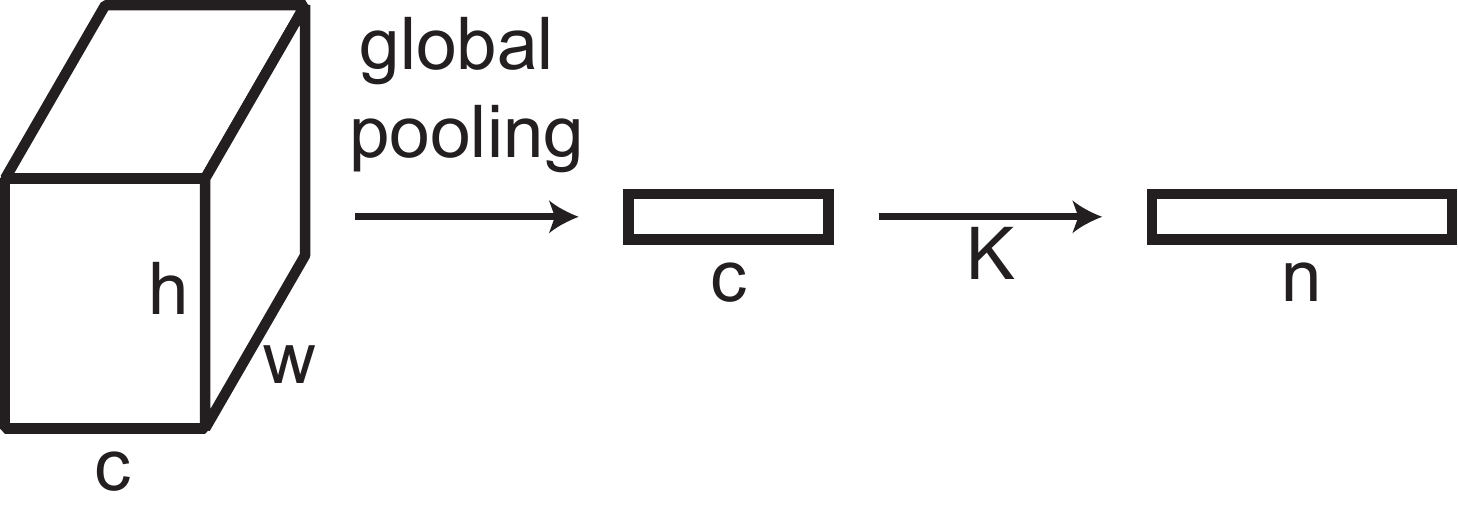}
        \caption{Classification output}
    \end{subfigure}
    \vspace{3mm}\\
    \begin{subfigure}[b]{0.5\linewidth}
        \includegraphics[width=\textwidth]{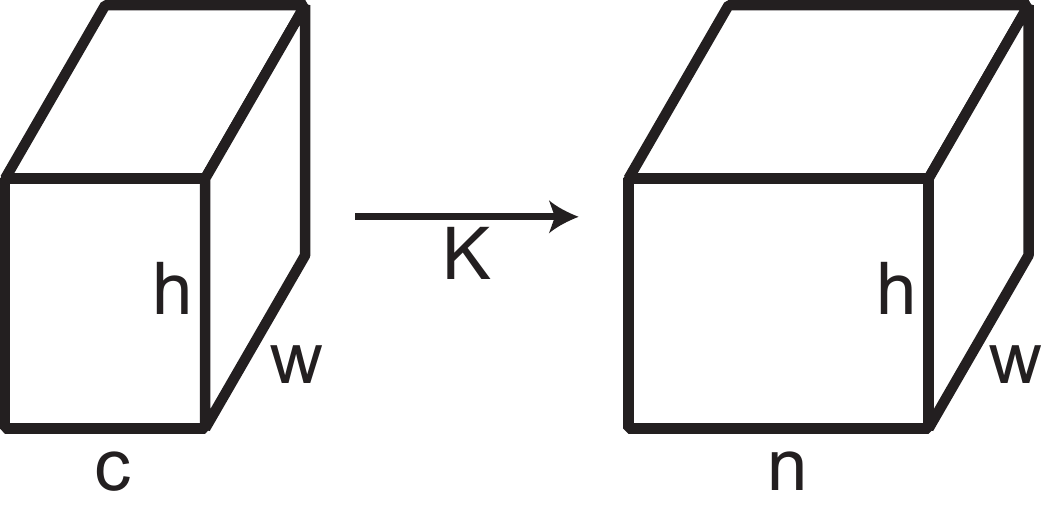}
        \caption{Localization output}
        \label{fig:vis_drn_after}
    \end{subfigure}
  \caption{Using a classification network for localization. The output stages of a DRN trained for image classification are shown in (a). Here $K$ is a $1\timess 1$ convolution that maps $c$ channels to $n$. To reconfigure the network for localization, we remove the pooling operator. The result is shown in (b). The reconfigured network produces $n$ activation maps of resolution $w \times h$. No training or parameter tuning is involved.}
  \label{fig:vis_drn}
\end{figure}

\begin{figure*}[t]
\centering

\begin{tabular}{@{}c @{\hskip 0.05in} c @{\hskip 0.05in} c @{\hskip 0.05in} c @{\hskip 0.05in} c@{}}

\includegraphics[width=0.19\linewidth]{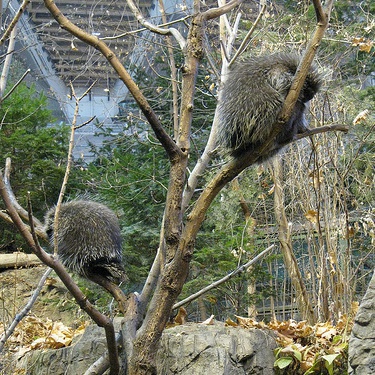}
&
\includegraphics[width=0.19\linewidth]{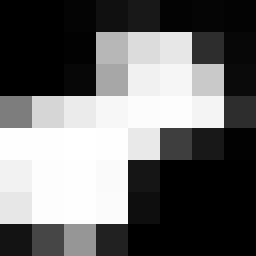}
&
\includegraphics[width=0.19\linewidth]{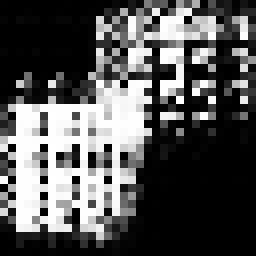}
&
\includegraphics[width=0.19\linewidth]{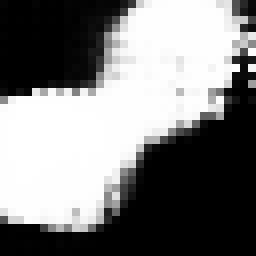}
&
\includegraphics[width=0.19\linewidth]{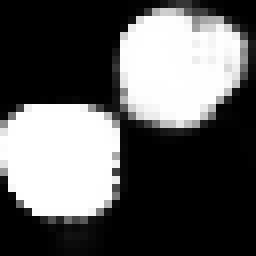}
 \\

\includegraphics[width=0.19\linewidth]{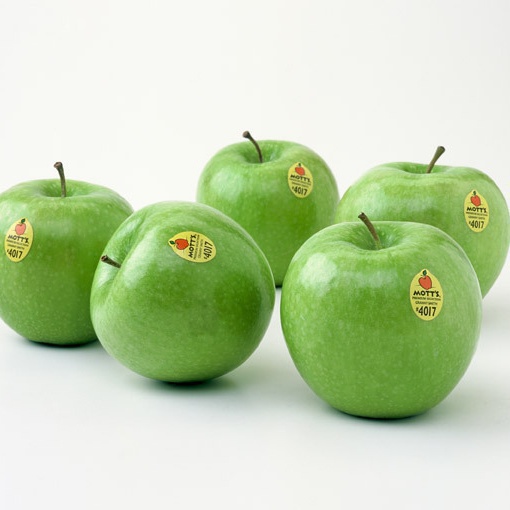}
&
\includegraphics[width=0.19\linewidth]{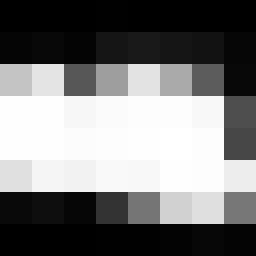}
&
\includegraphics[width=0.19\linewidth]{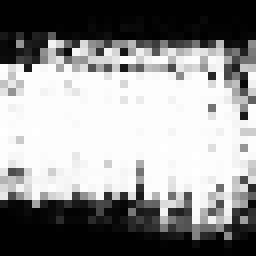}
&
\includegraphics[width=0.19\linewidth]{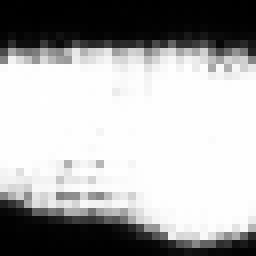}
&
\includegraphics[width=0.19\linewidth]{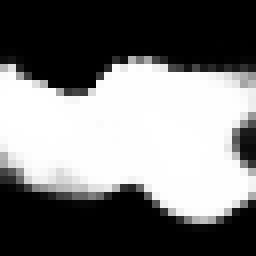}

\\

\includegraphics[width=0.19\linewidth]{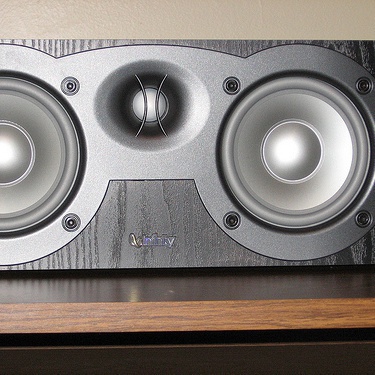}
&
\includegraphics[width=0.19\linewidth]{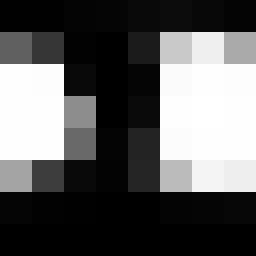}
&
\includegraphics[width=0.19\linewidth]{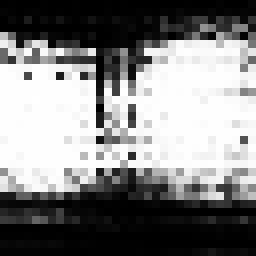}
&
\includegraphics[width=0.19\linewidth]{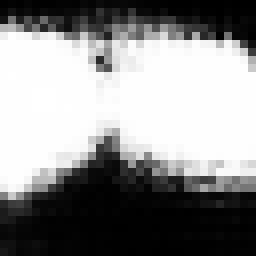}
&

\includegraphics[width=0.19\linewidth]{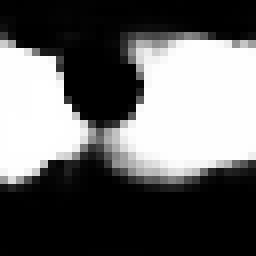}
 
\\




\includegraphics[width=0.19\linewidth]{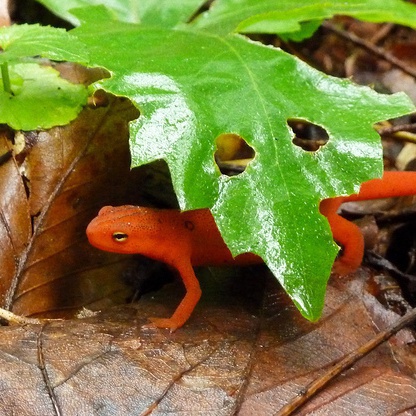}
&
\includegraphics[width=0.19\linewidth]{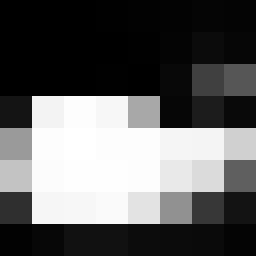}
&
\includegraphics[width=0.19\linewidth]{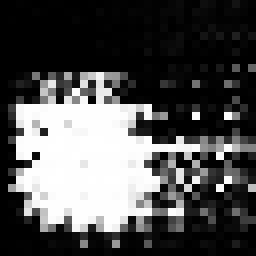}
&
\includegraphics[width=0.19\linewidth]{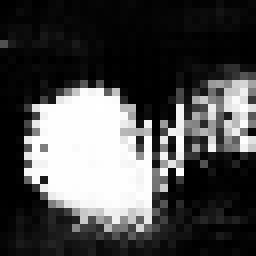}
&
\includegraphics[width=0.19\linewidth]{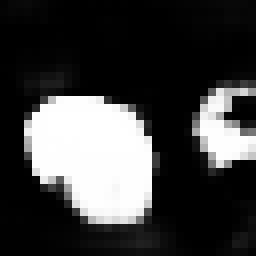}\\

(a) Input & (b) ResNet-18 & (c) DRN-A-18 & (d) DRN-B-26 & (e) DRN-C-26
\end{tabular}
\caption{Activation maps of ResNet-18 and corresponding DRNs. A DRN constructed from ResNet-18 as described in Section~\ref{sec:drn} is referred to as DRN-A-18. The corresponding DRN produced by the degridding scheme described in Section~\ref{sec:dive_drn} is referred to as DRN-C-26. The DRN-B-26 is an intermediate construction.}
\label{fig:activations}
\end{figure*}

To obtain high-resolution class activation maps, we remove the global average pooling operator. We then connect the final $1\timess 1$ convolution directly to $\G^5$. A softmax is applied to each column in the resulting volume to convert the pixelwise prediction scores to proper probability distributions. This procedure is illustrated in Figure~\ref{fig:vis_drn}.
The output of the resulting network is a set of activation maps that have the same spatial
resolution as $\G^5$ ($28\timess 28$). Each classification category $y$ has a corresponding activation map. For each pixel in this map, the map contains the probability that the object observed at this pixel is of category $y$.


The activation maps produced by our construction serve the same purpose as the results of the procedure of Zhou et al.~\cite{zhou2015learning}. However, the procedures are fundamentally different. Zhou et al.~worked with convolutional networks that produce drastically downsampled output that is not sufficiently resolved for object localization. For this reason, Zhou et al.~had to remove layers from the classification network, introduce parameters that compensate for the ablated layers, and then fine-tune the modified models to train the new parameters. Even then, the output resolution obtained by Zhou et al.~was quite small ($14\timess 14$) and the classification performance of the modified networks was impaired.

In contrast, the DRN was designed to produce high-resolution output maps and is trained in this configuration from the start. Thus the model trained for image classification already produces high-resolution activation maps. As our experiments will show, DRNs are more accurate than the original ResNets in image classification. Since DRNs produce high-resolution output maps from the start, there is no need to remove layers, add parameters, and retrain the model for localization. The original accurate classification model can be used for localization directly.

\section{Degridding}
\label{sec:dive_drn}

\begin{figure}[htb]
\centering
  \begin{tabular}{@{}c @{\hskip 2mm} c @{\hskip 2mm} c@{}}

\includegraphics[width=0.3\linewidth]{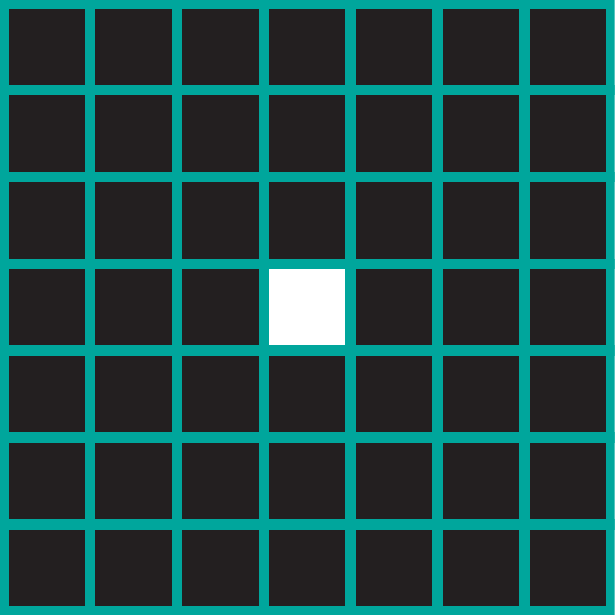}
&
\includegraphics[width=0.3\linewidth]{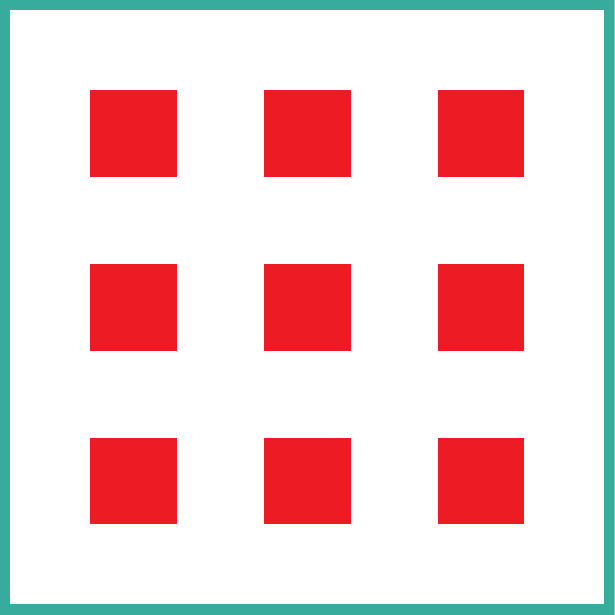}
&
\includegraphics[width=0.3\linewidth]{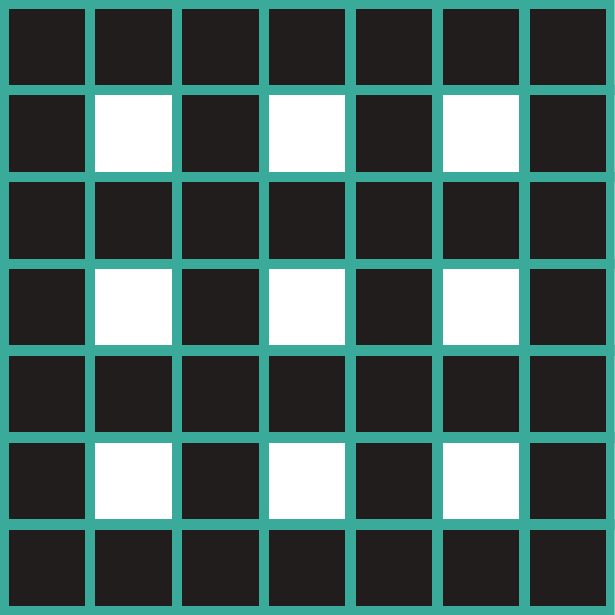}
\\
(a) Input & (b) Dilation 2 & (c) Output
\end{tabular}
\caption{A gridding artifact.}
\label{fig:gridding_toy}
\end{figure}

The use of dilated convolutions can cause gridding artifacts. Such artifacts are shown in Figure~\ref{fig:activations}(c) and have also been observed in concurrent work on semantic segmentation~\cite{wang2017understanding}.
Gridding artifacts occur when a feature map has higher-frequency content than the sampling rate of the dilated convolution. Figure~\ref{fig:gridding_toy} shows a didactic example. In Figure~\ref{fig:gridding_toy}(a), the input feature map has a single active pixel. A 2-dilated convolution (Figure~\ref{fig:gridding_toy}(b)) induces a corresponding grid pattern in the output (Figure~\ref{fig:gridding_toy}(c)).

\begin{figure*}[t]
\centering
\includegraphics[width=\linewidth]{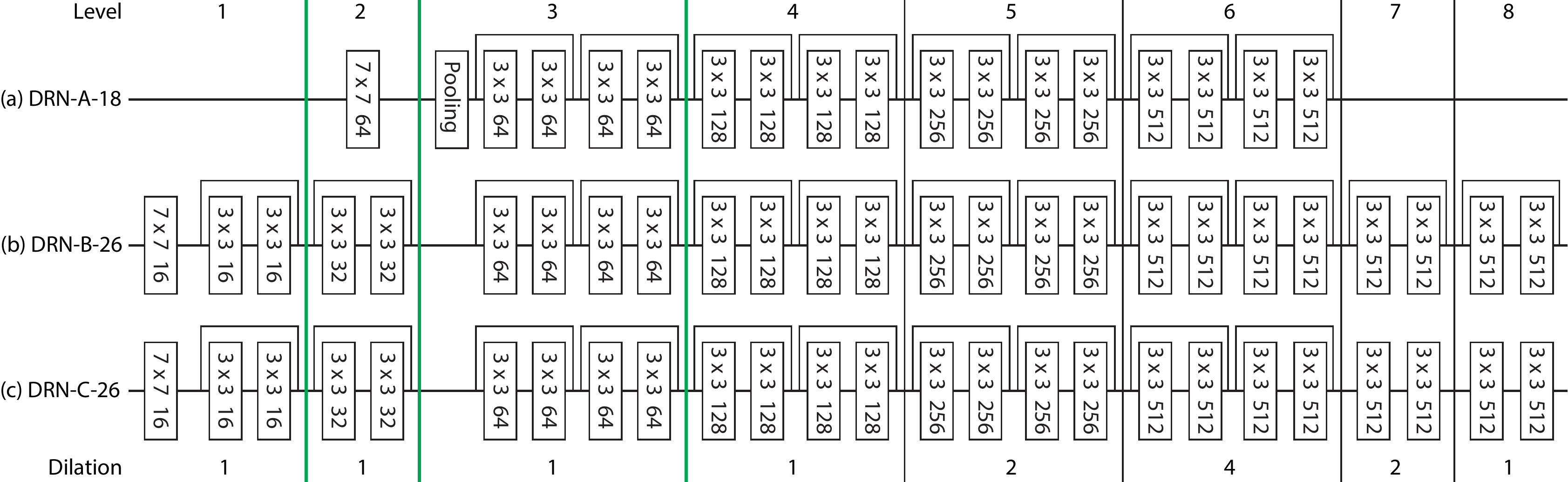}
\caption{Changing the DRN architecture to remove gridding artifacts from the output activation maps. Each rectangle is a Conv-BN-ReLU group and the numbers specify the filter size and the number of channels in that layer. The bold green lines represent downsampling by stride 2. The networks are divided into levels, such that all layers within a given level have the same dilation and spatial resolution. (a) DRN-A dilates the ResNet model directly, as described in Section~\ref{sec:drn}. (b) DRN-B replaces an early max pooling layer by residual blocks and adds residual blocks at the end of the network. (c) DRN-C removes residual connections from some of the added blocks. The rationale for each step is described in the text.}
\label{fig:drn_changes}
\end{figure*}

In this section, we develop a scheme for removing gridding artifacts from output activation maps produced by DRNs. The scheme is illustrated in Figure~\ref{fig:drn_changes}. A DRN constructed as described in Section~\ref{sec:drn} is referred to as DRN-A and is illustrated in Figure~\ref{fig:drn_changes}(a). An intermediate stage of the construction described in the present section is referred to as DRN-B and is illustrated in Figure~\ref{fig:drn_changes}(b). The final construction is referred to as DRN-C, illustrated in Figure~\ref{fig:drn_changes}(c).

\begin{figure}[h]
  \centering
  \begin{subfigure}[b]{0.325\linewidth}
        \includegraphics[width=\textwidth]{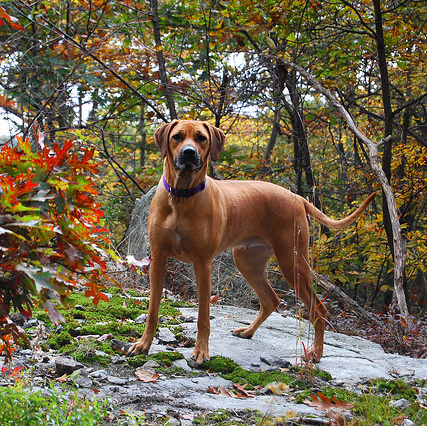}
        \caption{Input}
        \label{fig:max_pool_image}
    \end{subfigure}
    \begin{subfigure}[b]{0.325\linewidth}        \includegraphics[width=\textwidth]{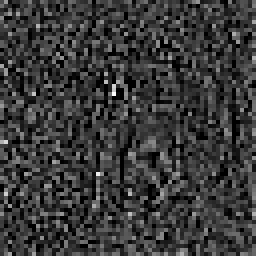}
        \caption{DRN-A-18}
        \label{fig:max_pool}
    \end{subfigure}
    \begin{subfigure}[b]{0.325\linewidth}
        \includegraphics[width=\textwidth]{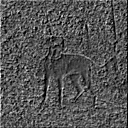}
        \caption{DRN-B-26}
        \label{fig:learned_pool}
    \end{subfigure}
    \vspace{-6mm}
  \caption{First stage of degridding, which modifies the early layers of the network. (b) and (c) show input feature maps for the first convolutional layer in level 3 of \mbox{DRN-A-18} and \mbox{DRN-B-26}. The feature map with the highest average activation is shown.}
\label{fig:max-pooling}
\end{figure}

\minisection{Removing max pooling.}
As shown in Figure~\ref{fig:drn_changes}(a), DRN-A inherits from the ResNet architecture a max pooling operation after the initial $7 \timess 7$ convolution.
We found that this max pooling operation leads to high-amplitude high-frequency activations, as shown in Figure~\ref{fig:max-pooling}(b).
Such high-frequency activations can be propagated to later layers and ultimately exacerbate gridding artifacts. We thus replace max pooling by convolutional filters, as shown in Figure~\ref{fig:drn_changes}(b). The effect of this transformation is shown in Figure~\ref{fig:max-pooling}(c).

\minisection{Adding layers.}
To remove gridding artifacts, we add convolutional layers at the end of the network, with progressively lower dilation. Specifically, after the last 4-dilated layer in DRN-A (Figure~\ref{fig:drn_changes}(a)), we add a 2-dilated residual block followed by a 1-dilated block. These become levels 7 and 8 in DRN-B, shown in Figure~\ref{fig:drn_changes}(b). This is akin to removing aliasing artifacts using filters with appropriate frequency~\cite{triggs2001empirical}.

\minisection{Removing residual connections.}
Adding layers with decreasing dilation, as described in the preceding paragraph, does not remove gridding artifacts entirely because of residual connections. The residual connections in levels 7 and 8 of DRN-B can propagate gridding artifacts from level 6. To remove gridding artifacts more effectively, we remove the residual connections in levels 7 and 8. This yields the DRN-C, our proposed construction, illustrated in Figure~\ref{fig:drn_changes}(c). Note that the \mbox{DRN-C} has higher depth and capacity than the corresponding \mbox{DRN-A} or the ResNet that had been used as the starting point. However, we will show that the presented degridding scheme has a dramatic effect on accuracy, such that the accuracy gain compensates for the added depth and capacity. For example, experiments will demonstrate that \mbox{DRN-C-26} has similar image classification accuracy to \mbox{DRN-A-34} and higher object localization and semantic segmentation accuracy than \mbox{DRN-A-50}.

\begin{figure*}[!ht]
  \centering

  \begin{tabular}{@{}c @{\hskip 0.07in} c @{\hskip 0.02in} c @{\hskip 0.02in} c @{\hskip 0.02in} c @{\hskip 0.02in} c @{\hskip 0.02in} c @{\hskip 0.05in} c@{}}

\includegraphics[width=0.12\linewidth]{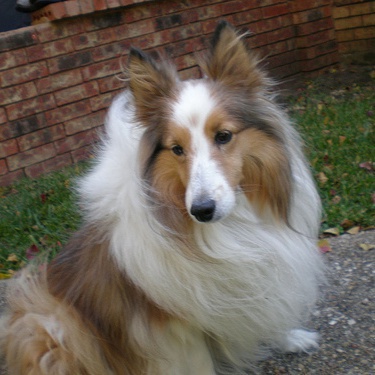}
&
\includegraphics[width=0.12\linewidth]{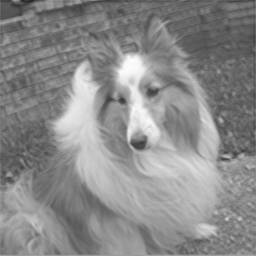}
&
\includegraphics[width=0.12\linewidth]{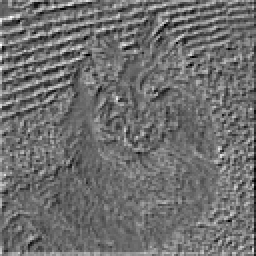}
&

\includegraphics[width=0.12\linewidth]{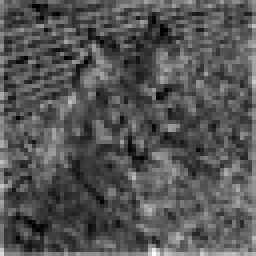}
&

\includegraphics[width=0.12\linewidth]{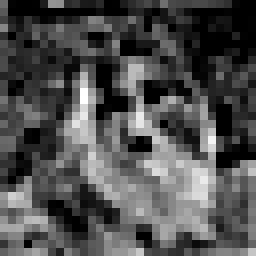}
&
\includegraphics[width=0.12\linewidth]{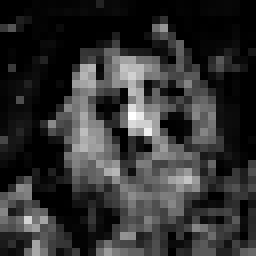}
&
\includegraphics[width=0.12\linewidth]{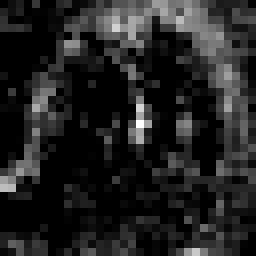}
&
\includegraphics[width=0.12\linewidth]{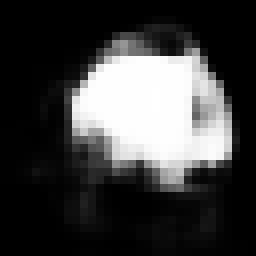}\\

\includegraphics[width=0.12\linewidth]{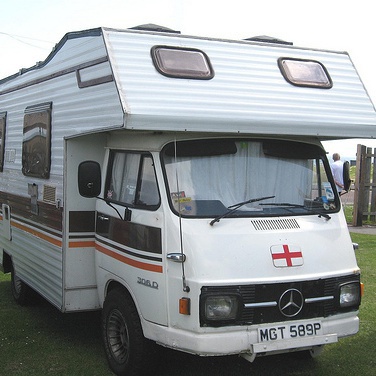}
&
\includegraphics[width=0.12\linewidth]{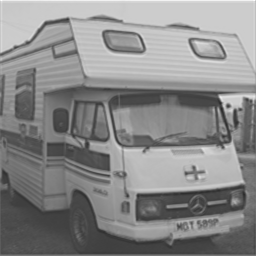}
&
\includegraphics[width=0.12\linewidth]{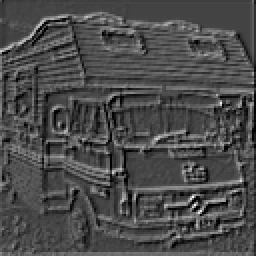}
&

\includegraphics[width=0.12\linewidth]{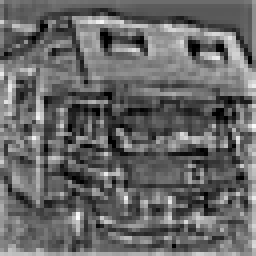}
&

\includegraphics[width=0.12\linewidth]{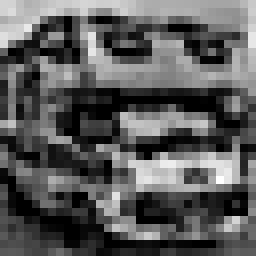}
&
\includegraphics[width=0.12\linewidth]{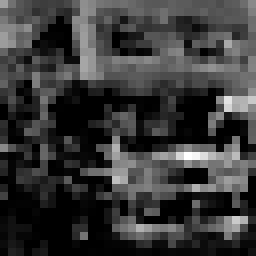}
&
\includegraphics[width=0.12\linewidth]{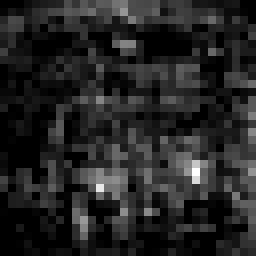}
&
\includegraphics[width=0.12\linewidth]{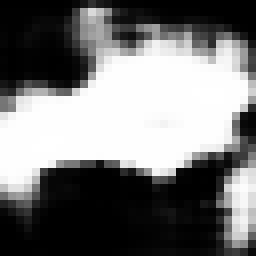}\\

\includegraphics[width=0.12\linewidth]{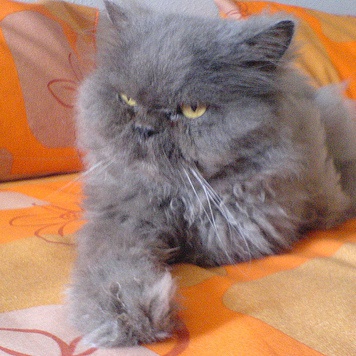}
&
\includegraphics[width=0.12\linewidth]{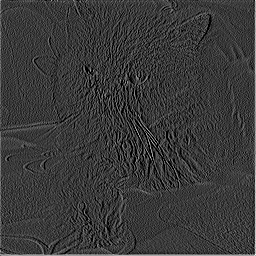}
&
\includegraphics[width=0.12\linewidth]{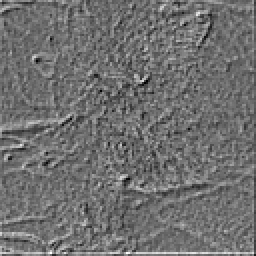}
&

\includegraphics[width=0.12\linewidth]{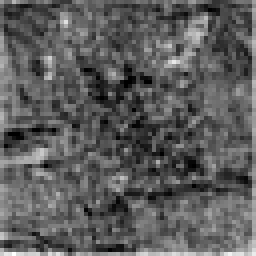}
&

\includegraphics[width=0.12\linewidth]{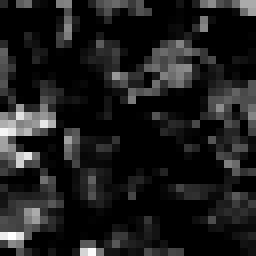}
&
\includegraphics[width=0.12\linewidth]{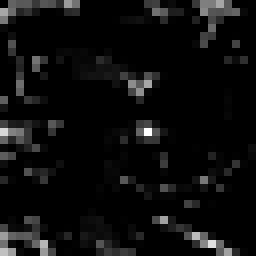}
&
\includegraphics[width=0.12\linewidth]{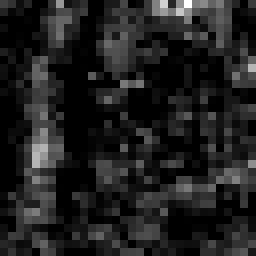}
&
\includegraphics[width=0.12\linewidth]{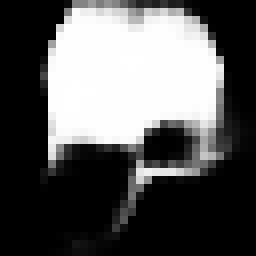}\\

Image & Level 1 & Level 2 & Level 3 & Level 4 & Level 5 & Level 6 & Class activation

\end{tabular}
\vspace{-1mm}
\caption{Activations inside a trained DRN-C-26. For each level, we show the feature map with the highest average activation magnitude among feature maps in the level's output. The levels are defined in Figure~\ref{fig:drn_changes}.}
\label{fig:layer_activation}
\end{figure*}

The activations inside a DRN-C are illustrated in Figure~\ref{fig:layer_activation}. This figure shows a feature map from the output of each level in the network. The feature map with the largest average activation magnitude is shown.


\section{Experiments}
\label{sec:experiments}

\subsection{Image Classification}
\label{sec:experiments-classification}

Training is performed on the ImageNet 2012 training set~\cite{imagenet}. The training procedure is similar to He et al.~\cite{resnet}. We use scale and aspect ratio augmentation as in Szegedy et al.~\cite{googlenet} and color perturbation as in Krizhevsky et al.~\cite{alexnet} and Howard~\cite{howard2013some}. Training is performed by SGD with momentum 0.9 and weight decay $10^{-4}$. The learning rate is initially set to $10^{-1}$ and is reduced by a factor of 10 every 30 epochs. Training proceeds for 120 epochs total.

The performance of trained models is evaluated on the ImageNet 2012
validation set. The images are resized so that the shorter side has 256 pixels. We use two evaluation protocols: 1-crop and 10-crop. In the 1-crop protocol, prediction accuracy is measured on the central $224 \timess 224$ crop. In the 10-crop protocol, prediction accuracy is measured on 10 crops from each image. Specifically, for each image we take the center crop, four corner crops, and flipped versions of these crops. The reported 10-crop accuracy is averaged over these 10 crops.

\minisection{ResNet vs.\ DRN-A.}
Table~\ref{tab:imagenet_comp} reports the accuracy of different models according to both evaluation protocols. Each DRN-A outperforms the corresponding ResNet model, despite having the same depth and capacity. For example, \mbox{DRN-A-18} and \mbox{DRN-A-34} outperform \mbox{ResNet-18} and \mbox{ResNet-34} in \mbox{1-crop} \mbox{top-1} accuracy by 2.43 and 2.92 percentage points, respectively. (A 10.5\% error reduction in the case of \mbox{ResNet-34 $\rightarrow$ DRN-A-34}.)

DRN-A-50 outperforms ResNet-50 in 1-crop top-1 accuracy by more than a percentage point. For comparison, the corresponding error reduction achieved by \mbox{ResNet-152} over \mbox{ResNet-101} is 0.3 percentage points. (From 22.44 to 22.16 on the center crop.)
These results indicate that even the direct transformation of a ResNet into a DRN-A, which does not change the depth or capacity of the model at all, significantly improves classification accuracy.

\begin{table}[htb]
\centering
\setlength{\tabcolsep}{1.75mm}
\ra{1.075}
\resizebox{1\linewidth}{!}{
\small
    \begin{tabular}{@{} l c cc c cc c c @{}}
    \toprule
    \multirow{2}[3]{*}{Model} && \multicolumn{2}{c}{1 crop} && \multicolumn{2}{c}{10 crops} && \multirow{2}[3]{*}{$P$}\\
    \cmidrule(lr){3-4} \cmidrule(lr){6-7}
     && top-1 & top-5  && top-1 & top-5 && \\
    \midrule
    ResNet-18 && 30.43 & 10.76 && 28.22 & 9.42 && 11.7M \\
    DRN-A-18 && 28.00 & 9.50 && 25.75 & 8.25 && 11.7M \\
    DRN-B-26 && 25.19 & 7.91 && 23.33 & 6.69 && 21.1M \\
    DRN-C-26 && 24.86 & 7.55 && 22.93 & 6.39 && 21.1M \\
    \midrule
    ResNet-34 && 27.73 & 8.74 && 24.76 & 7.35 && 21.8M \\
    DRN-A-34 && 24.81 & 7.54 && 22.64 & 6.34 && 21.8M \\
    DRN-C-42 && 22.94 & 6.57 && 21.20 & 5.60 && 31.2M \\
    \midrule
    ResNet-50 && 24.01 & 7.02 && 22.24 & 6.08 && 25.6M \\
    DRN-A-50 && 22.94 & 6.57 && 21.34 & 5.74 && 25.6M \\
    \midrule
    ResNet-101 && 22.44 & 6.21 && 21.08 & 5.35 && 44.5M \\
    \bottomrule
  \end{tabular}
}
\vspace{-1mm}
\caption{Image classification accuracy (error rates) on the ImageNet 2012 validation set. Lower is better. $P$ is the number of parameters in each model.}
\label{tab:imagenet_comp}
\end{table}

\minisection{DRN-A vs.\ DRN-C.}
Table~\ref{tab:imagenet_comp} also shows that the degridding construction described in Section~\ref{sec:dive_drn} is beneficial. Specifically, each \mbox{DRN-C} significantly outperforms the corresponding \mbox{DRN-A}. Although the degridding procedure increases depth and capacity, the resultant increase in accuracy is so substantial that the transformed DRN matches the accuracy of deeper models. Specifically, \mbox{DRN-C-26}, which is derived from \mbox{DRN-A-18}, matches the accuracy of the deeper \mbox{DRN-A-34}. In turn, \mbox{DRN-C-42}, which is derived from \mbox{DRN-A-34}, matches the accuracy of the deeper \mbox{DRN-A-50}. Comparing the degridded DRN to the original ResNet models, we see that \mbox{DRN-C-42} approaches the accuracy of \mbox{ResNet-101}, although the latter is deeper by a factor of 2.4.

\subsection{Object Localization}
\label{sec:experiments-localization}

We now evaluate the use of DRNs for weakly-supervised object localization, as described in Section~\ref{sec:vis}. As shown in Figure~\ref{fig:activations}, class activation maps provided by DRNs are much better spatially resolved than activation maps extracted from the corresponding ResNet.

We evaluate the utility of the high-resolution activation maps provided by DRNs for weakly-supervised object localization using the ImageNet 2012 validation set. We first predict the image categories based on 10-crop testing. Since the ground truth is in the form of bounding boxes, we need to fit bounding boxes to the activation maps.
We predict the object bounding boxes by analyzing the class responses on all the response maps. The general idea is to find tight bounding boxes that cover pixels for which the dominant response indicates the correct object class. Specifically, given
$\CC$ response maps of resolution $\WW \timess \HH$, let
$\ff(c, w, h)$ be the response at location $(w, h)$ on the $c^{\text{th}}$ response
map. In the ImageNet dataset, $C$ is 1000. We identify the dominant class at each location:
$$\gg(w, h) = \big\{c \ |\ \forall 1 \leq c^{\prime} \leq \CC.\ \ff(c, w, h) \geq \ff(c^{\prime}, w, h)\big\}.$$
For each class $c_i$, define the set of valid bounding boxes as
\begin{multline*}
\B_i = \big\{((w_1, h_1), (w_2, h_2)) | \\ \forall \gg(w, h) = c_i
\text{ and }
\ff(w, h, c_i) > t.\\ w_1 \leq
w \leq w_2 \text{ and } h_1 \leq h \leq h_2 \big\},
\end{multline*}
where $t$ is an activation threshold. The minimal bounding box for class $c_i$ is defined as
\begin{multline*}
\bb_i = \argmin_{((w_1, h_1), (w_2, h_2))\in \B_i } (w_2 - w_1) (h_2 -
h_1).
\end{multline*}

To evaluate the accuracy of DRNs on weakly-supervised object localization, we simply compute the minimal bounding box $\bb_i$ for the predicted class $i$ on each image. In the localization challenge, a predicted bounding box is considered accurate when its IoU with the ground-truth box is greater than 0.5. Table~\ref{tab:weak_location} reports the results. Note that the classification networks are used for localization directly, with no fine-tuning.

As shown in Table~\ref{tab:weak_location}, DRNs outperform the corresponding ResNet models. (Compare \mbox{ResNet-18} to \mbox{DRN-A-18}, \mbox{ResNet-34} to \mbox{DRN-A-34}, and \mbox{ResNet-50} to \mbox{DRN-A-50}.) This again illustrates the benefits of the basic DRN construction presented in Section~\ref{sec:drn}. Furthermore, \mbox{DRN-C-26} significantly outperforms \mbox{DRN-A-50}, despite having much lower depth. This indicates that that the degridding scheme described in Section~\ref{sec:dive_drn} has particularly significant benefits for applications that require more detailed spatial image analysis. \mbox{DRN-C-26} also outperforms \mbox{ResNet-101}.


\begin{table}[htb]
\centering
\ra{1.1}
  \begin{tabular}{@{} l @{\hspace{10mm}} c @{\hspace{6mm}} c @{}}
  \toprule
    Model & top-1 & top-5 \\
    \midrule
    ResNet-18 & 61.5 & 59.3 \\
    DRN-A-18 & 54.6 & 48.2 \\
    DRN-B-26 & 53.8 & 49.3 \\
    DRN-C-26 & 52.3 & 47.7 \\
    \midrule
    ResNet-34 & 58.7 & 56.4 \\
    DRN-A-34 & 55.5 & 50.7 \\
    DRN-C-42 & 50.7 & 46.8 \\
    \midrule
    ResNet-50 & 55.7 & 52.8 \\
    DRN-A-50 & 54.0 & 48.4 \\
    \midrule
    ResNet-101 & 54.6 & 51.9 \\
    \bottomrule
  \end{tabular}
\caption{Weakly-supervised object localization error rates on the ImageNet validation set. Lower is better. The degridded \mbox{DRN-C-26} outperforms \mbox{DRN-A-50}, despite lower depth and classification accuracy. \mbox{DRN-C-26} also outperforms \mbox{ResNet-101}.}
\label{tab:weak_location}
\end{table}








\begin{table*}[htbp]
\centering
\ra{1.2}
\resizebox{1\linewidth}{!}{
\small
  \begin{tabular}{@{}l@{\hspace{5mm}}*{19}{@{\hspace{2mm}}c}@{\hspace{5mm}}c@{}}
  &  \ver{Road} & \ver{Sidewalk} & \ver{Building} & \ver{Wall} &
    \ver{Fence} & \ver{Pole} & \ver{Light} & \ver{Sign} & \ver{Vegetation} & \ver{Terrain} & \ver{Sky} & \ver{Person} & \ver{Rider}
    & \ver{Car} & \ver{Truck} & \ver{Bus} & \ver{Train} &
    \ver{Motorcycle} & \ver{Bicycle} & \ver{mean IoU} \\
  \midrule
    DRN-A-50 & 96.9 & 77.4 & 90.3 & 35.8 & 42.8 & 59.0 & {\bf 66.8} & 74.5 & 91.6 & 57.0 & 93.4 & 78.7 & 55.3 & 92.1 & 43.2 & 59.5 & 36.2 & {\bf 52.0} & {\bf 75.2} & 67.3 \\
    DRN-C-26 & 97.4 & 80.7 & 90.4 & 36.1 & 47.0 & 56.9 & 63.8 & 73.0 & 91.2 &
    {\bf 57.9} & 93.4 & 77.3 & 53.8 & 92.7 & 45.0 & 70.5 & 48.4 & 44.2 & 72.8
    & 68.0 \\
    DRN-C-42 & {\bf 97.7} & {\bf 82.2} & {\bf 91.2} & {\bf 40.5} & {\bf 52.6} & {\bf 59.2} & 66.7 & {\bf 74.6} & {\bf 91.7} & 57.7 & {\bf 94.1} & {\bf 79.1} & {\bf 56.0} & {\bf 93.6} & {\bf 56.0} & {\bf 74.3} & {\bf 54.7} & 50.9 & 74.1 & {\bf 70.9} \\
  \midrule
  \end{tabular}
}
\vspace{-1mm}
  \caption{Performance of dilated residual networks on the Cityscapes validation set. Higher is better. DRN-C-26 outperforms DRN-A-50, despite lower depth. DRN-C-42 achieves even higher accuracy. For reference, a comparable baseline setup of ResNet-101 was reported to achieve a mean IoU of 66.6.}
\label{tab:cityscapes}
\end{table*}

\begin{figure*}[htb]
\centering
  \begin{tabular}{@{}c @{\hspace{1mm}} c  @{\hspace{1mm}} c @{\hspace{1mm}} c@{}}

\includegraphics[width=0.24\linewidth]{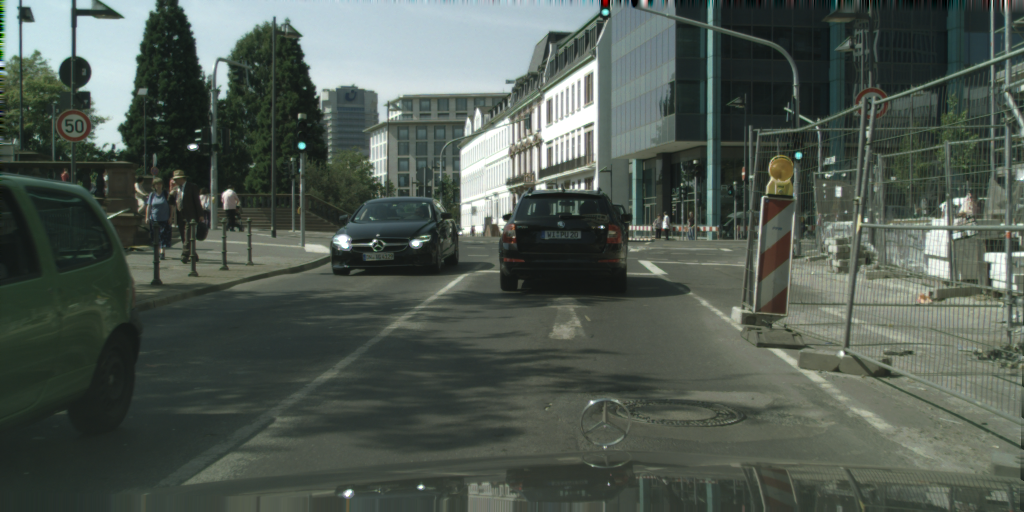}
&
\includegraphics[width=0.24\linewidth]{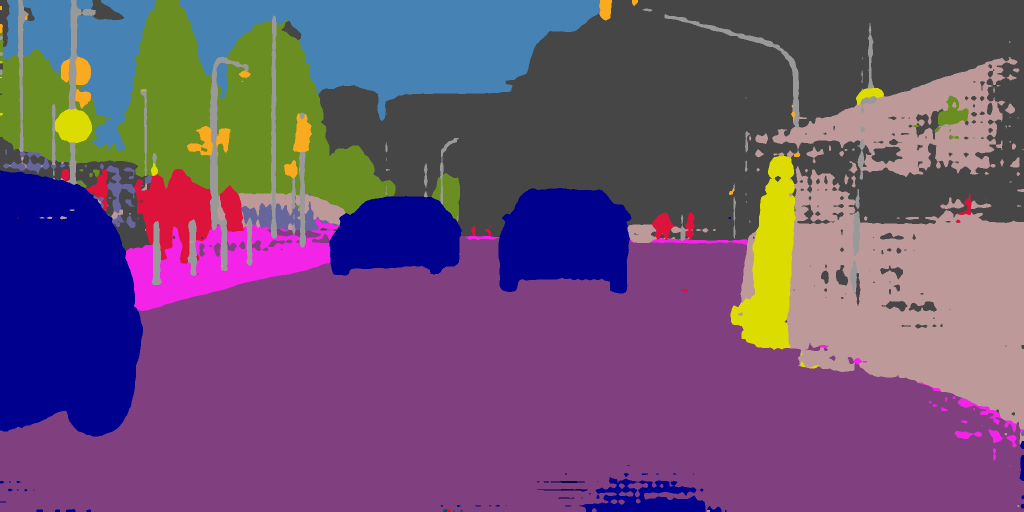}
&
\includegraphics[width=0.24\linewidth]{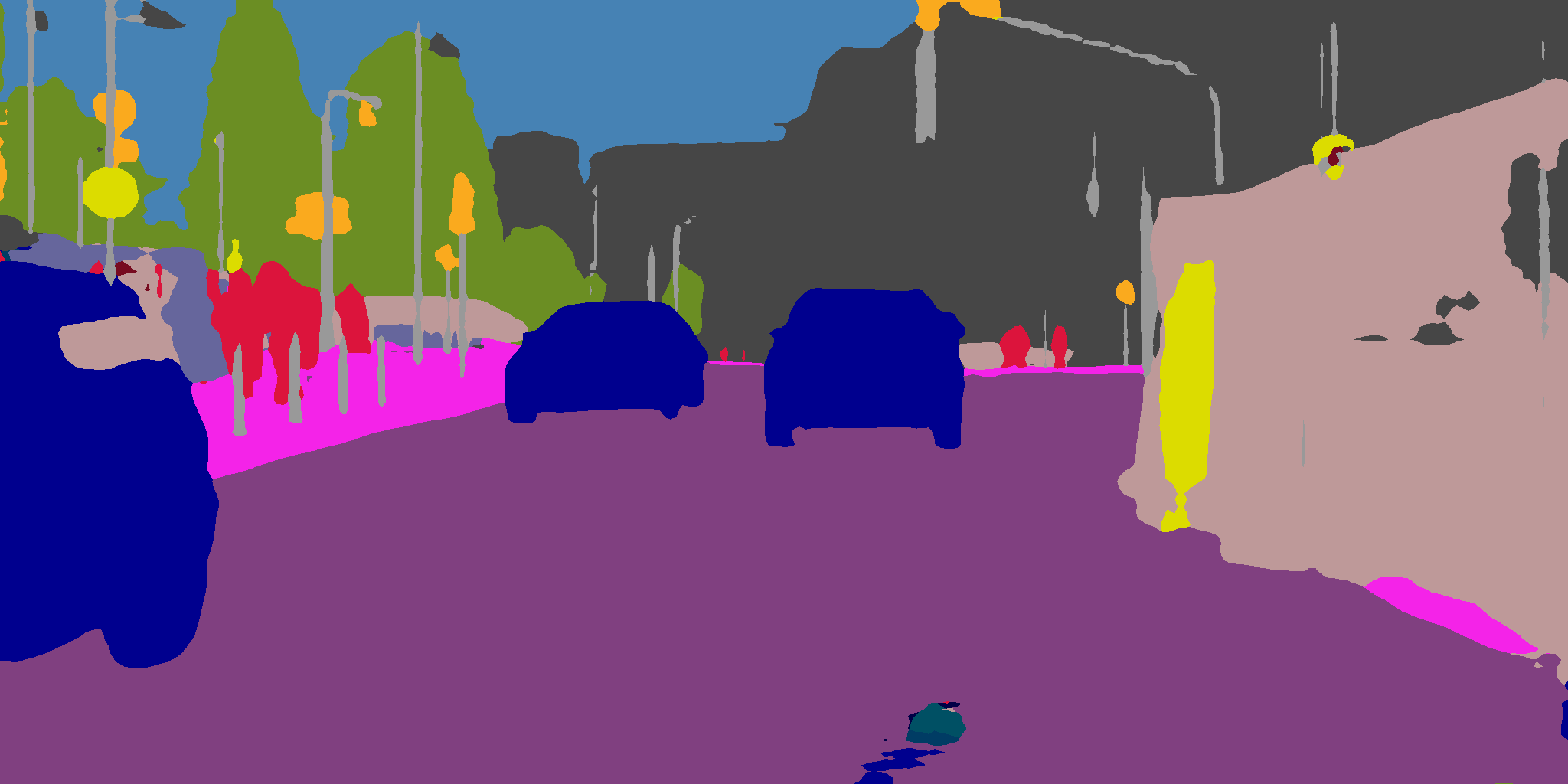}
&
\includegraphics[width=0.24\linewidth]{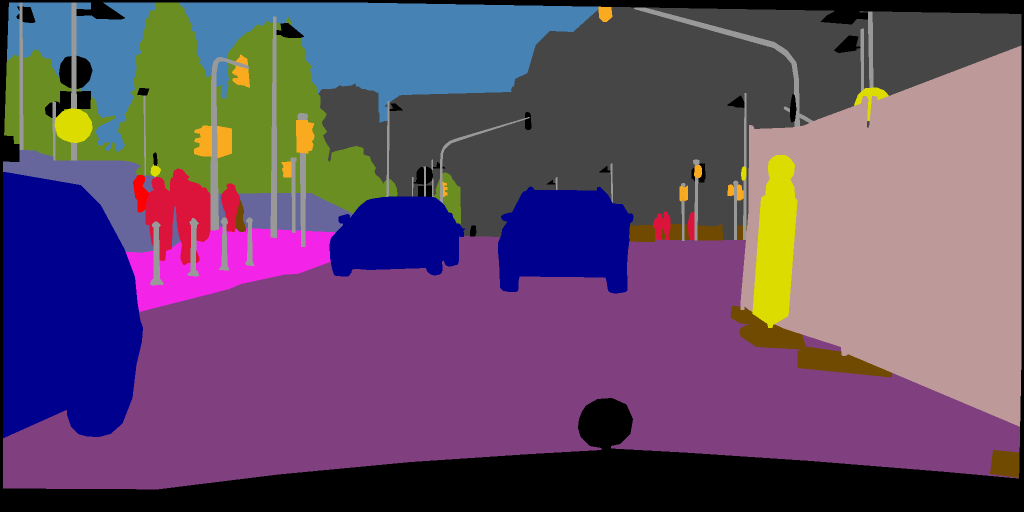}
\\

\includegraphics[width=0.24\linewidth]{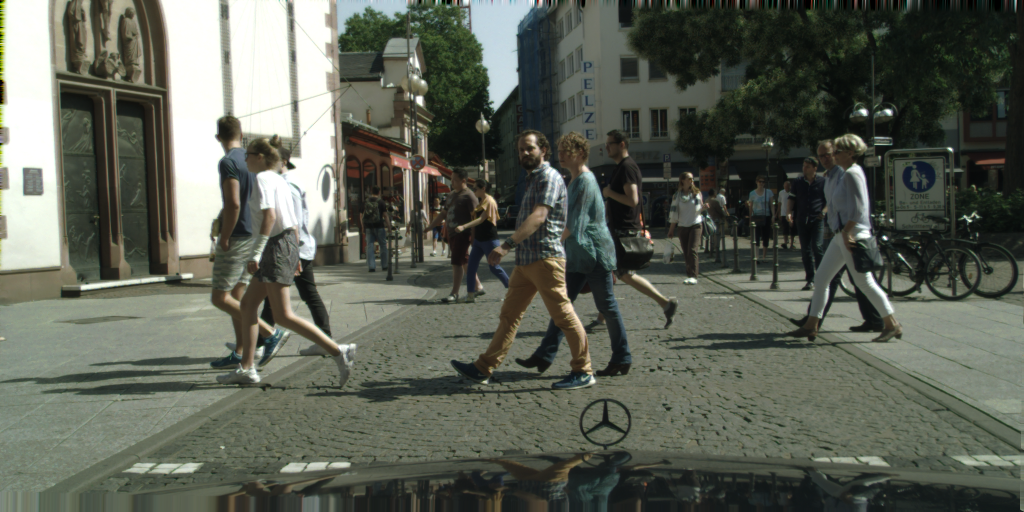}
&
\includegraphics[width=0.24\linewidth]{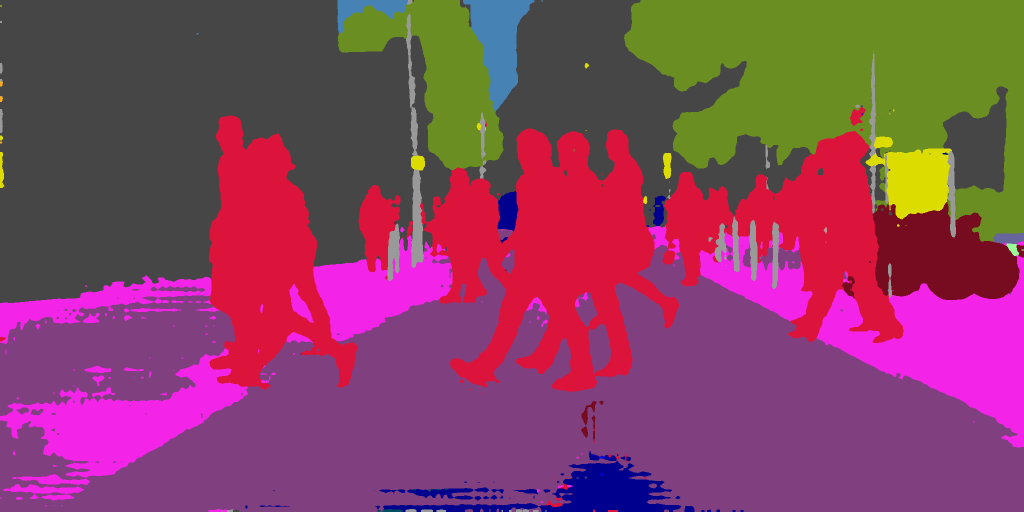}
&
\includegraphics[width=0.24\linewidth]{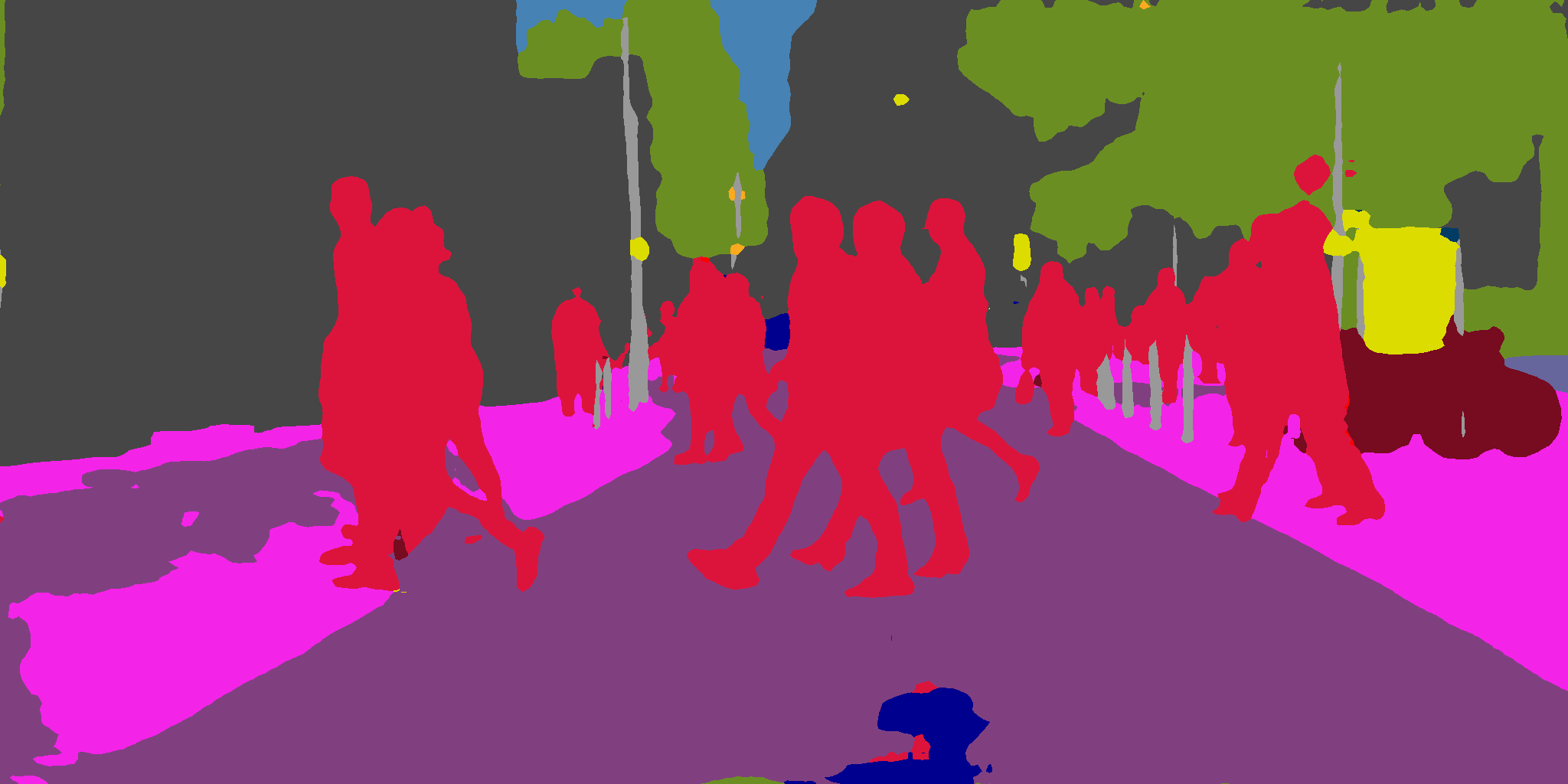}
&
\includegraphics[width=0.24\linewidth]{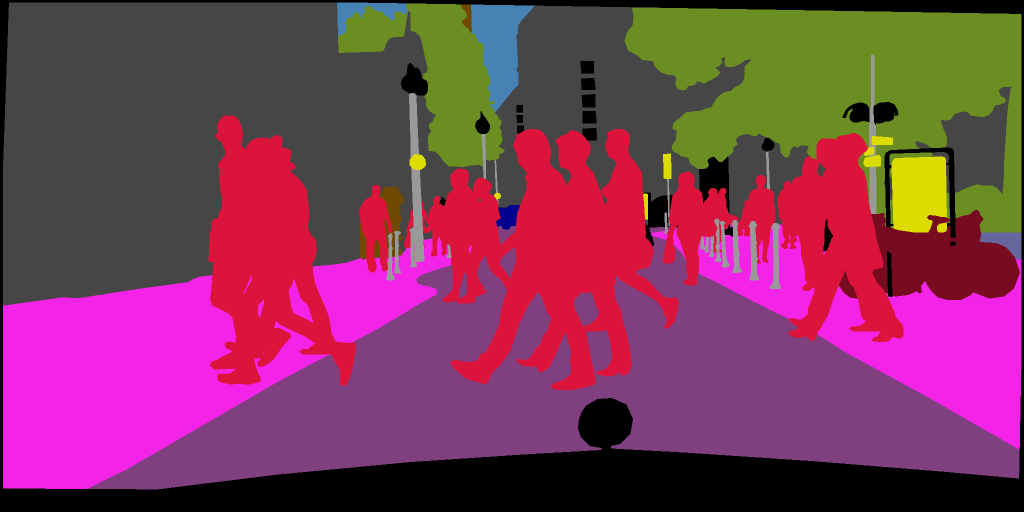}
\\

\includegraphics[width=0.24\linewidth]{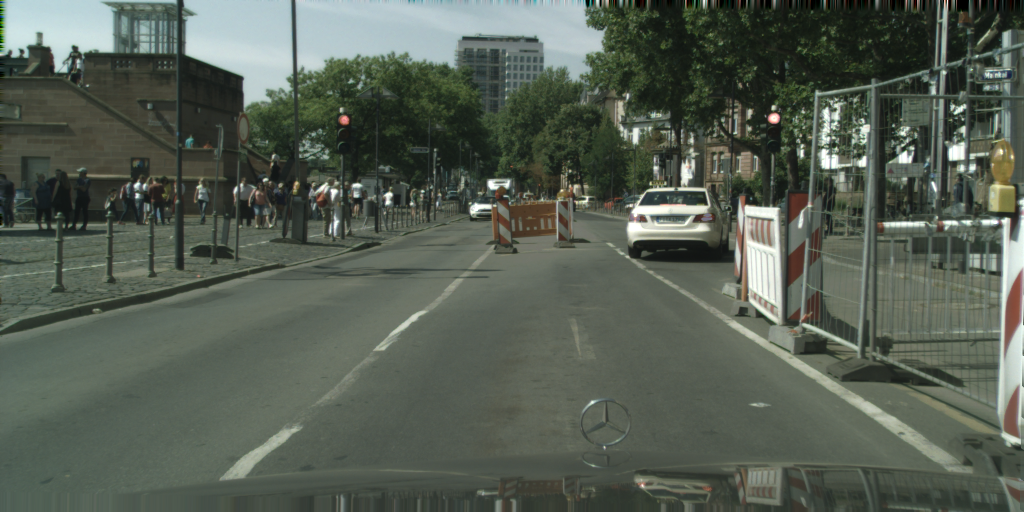}
&
\includegraphics[width=0.24\linewidth]{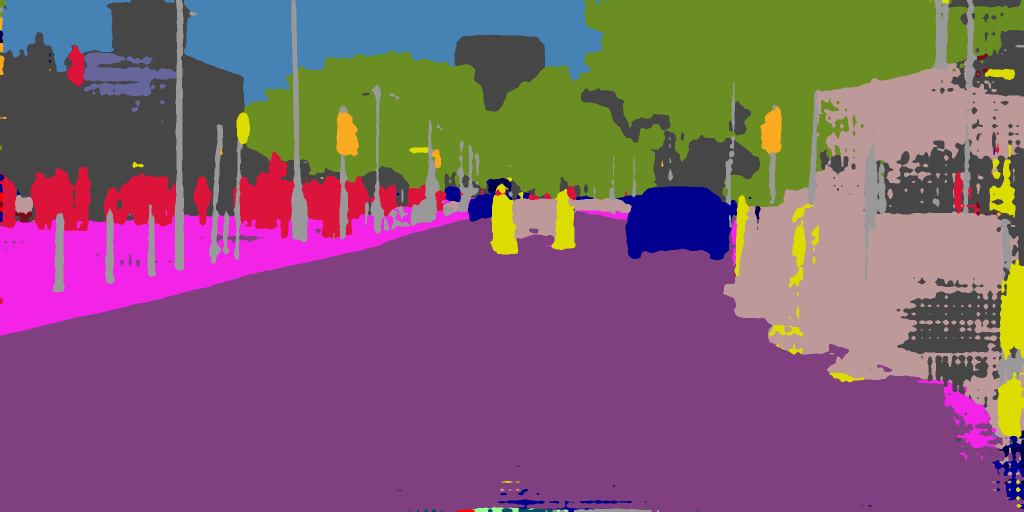}
&
\includegraphics[width=0.24\linewidth]{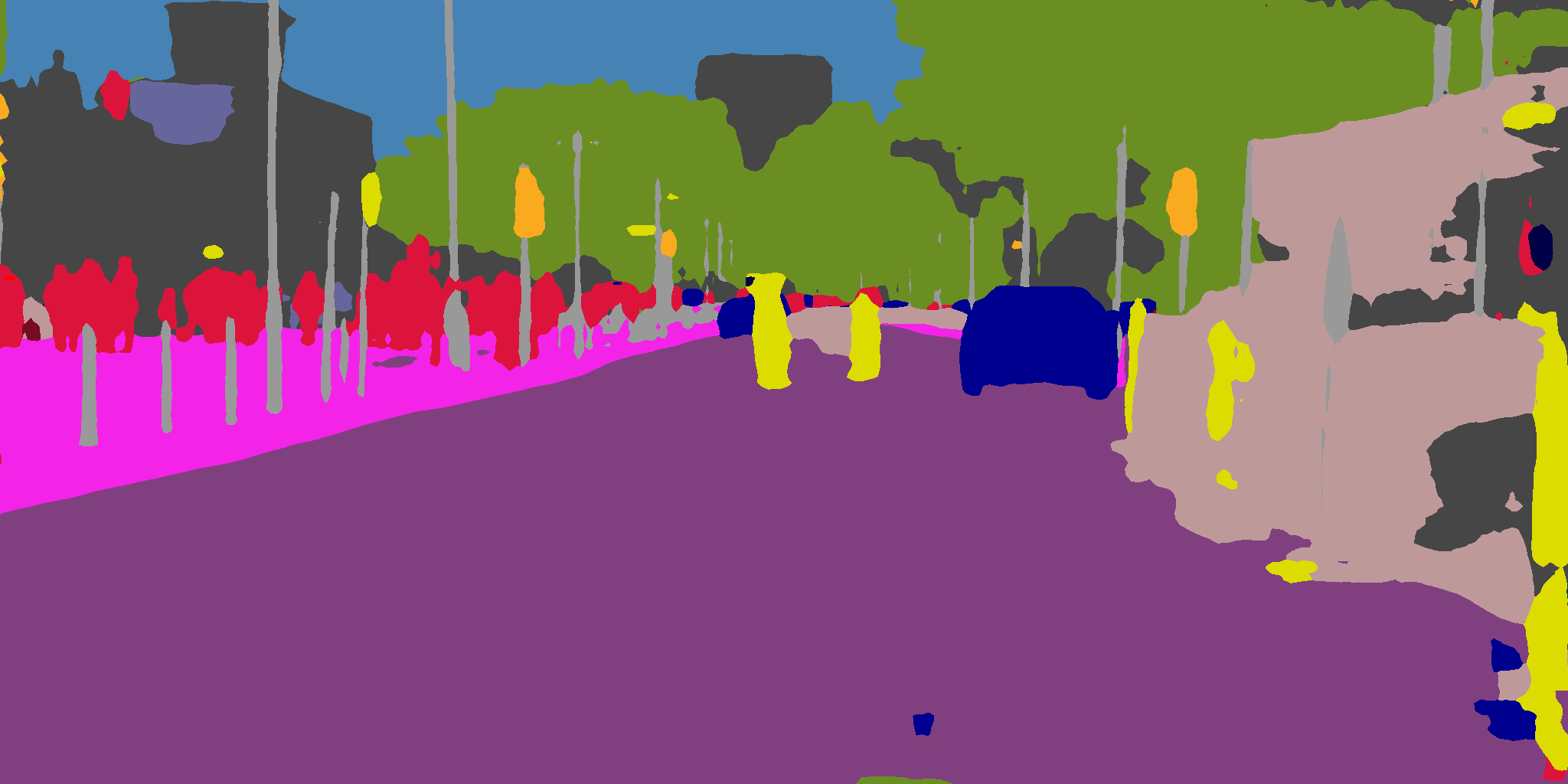}
&
\includegraphics[width=0.24\linewidth]{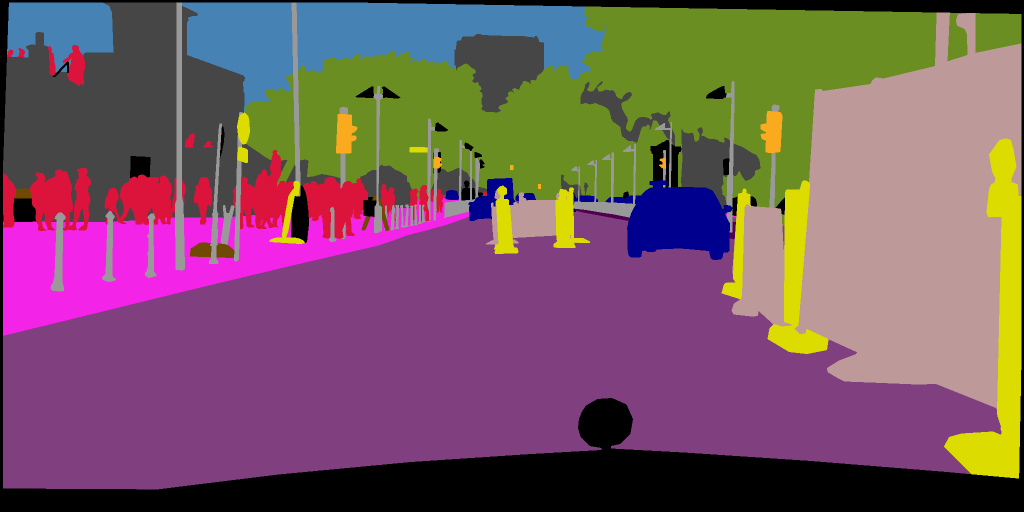}
\\

(a) Input & (b) DRN-A-50 & (c) DRN-C-26 & (d) Ground truth
\end{tabular}
\caption{Semantic segmentation on the Cityscapes dataset. The degridded DRN-C-26 produces cleaner results than the deeper DRN-A-50.}
\label{fig:cityscapes}
\end{figure*}

\subsection{Semantic Segmentation}
\label{sec:experiments-segmentation}

We now transfer DRNs to semantic segmentation. High-resolution internal representations are known to be important for this task~\cite{fcn,dilation,Cordts2016}. Due to the severe downsampling in prior image classification architectures, their transfer to semantic segmentation necessitated post-hoc adaptations such as up-convolutions, skip connections, and post-hoc dilation~\cite{fcn,chen2016deeplab,Noh2015,dilation}. In contrast, the high resolution of the output layer in a DRN means that we can transfer a classification-trained DRN to semantic segmentation by simply removing the global pooling layer and operating the network fully-convolutionally~\cite{fcn}, without any additional structural changes. The predictions synthesized by the output layer are upsampled to full resolution using bilinear interpolation, which does not involve any parameters.

We evaluate this capability using the Cityscapes dataset~\cite{Cordts2016}. We use the standard Cityscapes training and validation sets. To understand the properties of the models themselves, we only use image cropping and mirroring for training. We do not use any other data augmentation and do not append additional modules to the network. The results are reported in Table~\ref{tab:cityscapes}.

All presented models outperform a comparable baseline setup of ResNet-101, which was reported to achieve a mean IoU of 66.6~\cite{chen2016deeplab}. For example, \mbox{DRN-C-26} outperforms the ResNet-101 baseline by more than a percentage point, despite having 4 times lower depth. The \mbox{DRN-C-42} model outperforms the ResNet-101 baseline by more than 4 percentage points, despite 2.4 times lower depth.

Comparing different DRN models, we see that both \mbox{DRN-C-26} and \mbox{DRN-C-42} outperform \mbox{DRN-A-50}, suggesting that the degridding construction presented in Section~\ref{sec:dive_drn} is particularly beneficial for dense prediction tasks. A qualitative comparison between \mbox{DRN-A-50} and \mbox{DRN-C-26} is shown in Figure~\ref{fig:cityscapes}. As the images show, the predictions of \mbox{DRN-A-50} are marred by gridding artifacts even though the model was trained with dense pixel-level supervision. In contrast, the predictions of DRN-C-26 are not only more accurate, but also visibly cleaner.

\section{Conclusion}

We have presented an approach to designing convolutional networks for image analysis. Rather than progressively reducing the resolution of internal representations until the spatial structure of the scene is no longer discernible, we keep high spatial resolution all the way through the final output layers. We have shown that this simple transformation improves image classification accuracy, outperforming state-of-the-art models. We have then shown that accuracy can be increased further by modifying the construction to alleviate gridding artifacts introduced by dilation.

The presented image classification networks produce informative output activations, which can be used directly for weakly-supervised object localization, without any fine-tuning. The presented models can also be used for dense prediction tasks such as semantic segmentation, where they outperform deeper and higher-capacity baselines.

The results indicate that dilated residual networks can be used as a starting point for image analysis tasks that involve complex natural images, particularly when detailed understanding of the scene is important. We will release code and pretrained models to support future research and applications.

\section*{Acknowledgments}
This work was supported by Intel and the National Science Foundation (IIS-1251217 and VEC 1539014/1539099).

\clearpage

\balance

{\small
\bibliographystyle{ieee}
\bibliography{drn}

\begin{thebibliography}{10}\itemsep=-1pt

\bibitem{chen2016deeplab}
L.-C. Chen, G.~Papandreou, I.~Kokkinos, K.~Murphy, and A.~L. Yuille.
\newblock {DeepLab}: Semantic image segmentation with deep convolutional nets,
  atrous convolution, and fully connected {CRFs}.
\newblock {\em arXiv:1606.00915}, 2016.

\bibitem{Cordts2016}
M.~Cordts, M.~Omran, S.~Ramos, T.~Rehfeld, M.~Enzweiler, R.~Benenson,
  U.~Franke, S.~Roth, and B.~Schiele.
\newblock The {Cityscapes} dataset for semantic urban scene understanding.
\newblock In {\em CVPR}, 2016.

\bibitem{GalleguillosBelongie2010}
C.~Galleguillos and S.~J. Belongie.
\newblock Context based object categorization: A critical survey.
\newblock {\em Computer Vision and Image Understanding}, 114(6), 2010.

\bibitem{rcnn}
R.~B. Girshick, J.~Donahue, T.~Darrell, and J.~Malik.
\newblock Region-based convolutional networks for accurate object detection and
  segmentation.
\newblock {\em PAMI}, 38(1), 2016.

\bibitem{Hariharan2015}
B.~Hariharan, P.~A. Arbel{\'{a}}ez, R.~B. Girshick, and J.~Malik.
\newblock Hypercolumns for object segmentation and fine-grained localization.
\newblock In {\em CVPR}, 2015.

\bibitem{resnet}
K.~He, X.~Zhang, S.~Ren, and J.~Sun.
\newblock Deep residual learning for image recognition.
\newblock In {\em CVPR}, 2016.

\bibitem{howard2013some}
A.~G. Howard.
\newblock Some improvements on deep convolutional neural network based image
  classification.
\newblock {\em arXiv:1312.5402}, 2013.

\bibitem{alexnet}
A.~Krizhevsky, I.~Sutskever, and G.~E. Hinton.
\newblock {ImageNet} classification with deep convolutional neural networks.
\newblock In {\em NIPS}, 2012.

\bibitem{LeCun1989}
Y.~LeCun, B.~Boser, J.~S. Denker, D.~Henderson, R.~E. Howard, W.~Hubbard, and
  L.~D. Jackel.
\newblock Backpropagation applied to handwritten zip code recognition.
\newblock {\em Neural Computation}, 1(4), 1989.

\bibitem{fcn}
J.~Long, E.~Shelhamer, and T.~Darrell.
\newblock Fully convolutional networks for semantic segmentation.
\newblock In {\em CVPR}, 2015.

\bibitem{Noh2015}
H.~Noh, S.~Hong, and B.~Han.
\newblock Learning deconvolution network for semantic segmentation.
\newblock In {\em ICCV}, 2015.

\bibitem{imagenet}
O.~Russakovsky, J.~Deng, H.~Su, J.~Krause, S.~Satheesh, S.~Ma, Z.~Huang,
  A.~Karpathy, A.~Khosla, M.~S. Bernstein, A.~C. Berg, and F.~Li.
\newblock {ImageNet} large scale visual recognition challenge.
\newblock {\em IJCV}, 115(3), 2015.

\bibitem{vggnet}
K.~Simonyan and A.~Zisserman.
\newblock Very deep convolutional networks for large-scale image recognition.
\newblock In {\em ICLR}, 2015.

\bibitem{googlenet}
C.~Szegedy, W.~Liu, Y.~Jia, P.~Sermanet, S.~E. Reed, D.~Anguelov, D.~Erhan,
  V.~Vanhoucke, and A.~Rabinovich.
\newblock Going deeper with convolutions.
\newblock In {\em CVPR}, 2015.

\bibitem{Torralba2008}
A.~Torralba, R.~Fergus, and W.~T. Freeman.
\newblock 80 million tiny images: {A} large data set for nonparametric object
  and scene recognition.
\newblock {\em PAMI}, 30(11), 2008.

\bibitem{triggs2001empirical}
B.~Triggs.
\newblock Empirical filter estimation for subpixel interpolation and matching.
\newblock In {\em ICCV}, 2001.

\bibitem{wang2017understanding}
P.~Wang, P.~Chen, Y.~Yuan, D.~Liu, Z.~Huang, X.~Hou, and G.~Cottrell.
\newblock Understanding convolution for semantic segmentation.
\newblock {\em arXiv:1702.08502}, 2017.

\bibitem{dilation}
F.~Yu and V.~Koltun.
\newblock Multi-scale context aggregation by dilated convolutions.
\newblock In {\em ICLR}, 2016.

\bibitem{zhou2015learning}
B.~Zhou, A.~Khosla, {\`{A}}.~Lapedriza, A.~Oliva, and A.~Torralba.
\newblock Learning deep features for discriminative localization.
\newblock In {\em CVPR}, 2016.

\end{thebibliography}
}

\end{document}